# Explainable Data Poison Attacks on Human Emotion Evaluation Systems based on EEG Signals

Zhibo Zhang, Sani Umar, Ahmed Y. Al Hammadi, Sangyoung Yoon, Ernesto Damiani, *Senior Member, IEEE*, Claudio Agostino Ardagna, *Senior Member, IEEE*, Nicola Bena, and Chan Yeob Yeun, *Senior Member, IEEE*

*Abstract--* The major aim of this paper is to explain the data poisoning attacks using label-flipping during the training stage of the electroencephalogram (EEG) signal-based human emotion evaluation systems deploying Machine Learning models from the attackers' perspective. Human emotion evaluation using EEG signals has consistently attracted a lot of research attention. The identification of human emotional states based on EEG signals is effective to detect potential internal threats caused by insider individuals. Nevertheless, EEG signal-based human emotion evaluation systems have shown several vulnerabilities to data poison attacks. Besides, due to the instability and complexity of the EEG signals, it is challenging to explain and analyze how data poison attacks influence the decision process of EEG signal-based human emotion evaluation systems. In this paper, from the attackers' side, data poison attacks occurring in the training phases of six different Machine Learning models including Random Forest, Adaptive Boosting (AdaBoost), Extra Trees, XGBoost, Multilayer Perceptron (MLP), and K-Nearest Neighbors (KNN) intrude on the EEG signal-based human emotion evaluation systems using these Machine Learning models. This seeks to reduce the performance of the aforementioned Machine Learning models with regard to the classification task of 4 different human emotions using EEG signals. The findings of the experiments demonstrate that the suggested data poison assaults are model-independently successful, although various models exhibit varying levels of resilience to the attacks. In addition, the data poison attacks on the EEG signal-based human emotion evaluation systems are explained with several Explainable Artificial Intelligence (XAI) methods including Shapley Additive Explanation (SHAP) values, Local Interpretable Model-agnostic Explanations (LIME), and Generated Decision Trees. And the codes of this paper are publicly available on GitHub.

This work was supported in part by Technology Innovation Institute (TII) under Grant 8434000394. *(Corresponding author: Chan Yeob Yeun.)*

Zhibo Zhang, Sani Umar, Ahmed Y. Al Hammadi, Sangyoung Yoon, Ernesto Damiani, and Chan Yeob Yeun are with the Department of Electrical Engineering and Computer Science, Khalifa University, Abu Dhabi, United Arab Emirates. (email: 100060990@ku.ac.ae; 100059856@ku.ac.ae; ahmed.yalhammadi@ku.ac.ae; sangyoung.yoon@ku.ac.ae; ernesto.damiani@ku.ac.ae; chan.yeun@ku.ac.ae).

Claudio Agostino Ardagna and Nicola Bena are with Dipartimento di Informatica, Università degli Studi di Milano, Milano, Italy (e-mail: claudio.ardagna@unimi.it; nicola.bena@unimi.it).

## I. INTRODUCTION

HUMAN daily activities, such as communication, decision-making, and personal development, are significantly influenced by human emotions. Moreover, the unstable emotional states of people who are insiders in an industrial organization, such as current or former employees, can cause industrial insider risk [1]. Therefore, it is important to establish emotional recognition systems to detect the emotional fluctuations of industrial insider humans to avoid unnecessary industrial loss. The classification of human emotions has recently improved thanks to the ongoing development of artificial intelligence (AI) technologies, including deep learning and Machine Learning [2], along with cutting-edge therapeutic therapy.

Traditionally, facial expression [3] and voice [4] information were used to evaluate human emotions and actions of people. However, people are capable of easily masking facial and speech information with the right training [5]. On the other hand, since it is impossible for people to conceal or influence their brainwaves, EEG signals [6] have been used in recent years to evaluate a person's emotional state in order to stop possible industrial insider attacks. Moreover, several EEG signal-based human emotion evaluation systems have employed Machine Learning [7] classifiers in various configurations to analyze human emotions in the context of AI applications.

It has been demonstrated that it is possible to intentionally alter the training data to influence the Machine Learning models' decision-making process, which ultimately results in a complete breakdown during the testing (or inference) phase [8]. On the other hand, attackers could identify and exploit Machine Learning models' weaknesses [9] to reduce the effectiveness of EEG signal-based human emotion assessment systems. To achieve this, attackers employ data poisoning (DP) [10] attack techniques to taint a target Machine Learning model by poisoning it during the training phase. Therefore, the goal of this paper is to design explainable label-flipping-based DP attacks that specifically target the Machine Learning classifiers of EEG signal-based human emotion assessment systems. To quantify the poisoning impacts and vulnerabilities of each ML model, various poisoning thresholds have also been proposed in this research and Explainable Artificial Intelligence (XAI) [11] techniques are used to investigate and clarify the precise impact of DP attacks on EEG signal-based human emotion evaluation systems in terms of features and internal mechanisms.

Therefore, to fill the gap of deploying explainable label-flipping DP attacks against Machine Learning models of the EEG signal-based human emotion assessment systems from the attackers' perspective, the main contributions of this research paper are mentioned as follows:

1) This study covered diverse types of data manipulation and vulnerabilities of different Machine Learning models in the context of EEG signal-based human emotion evaluation systems.

2) This paper deployed two different DP categories of label-flipping attacks in the training stages of different Machine Learning models to test their resilience against different DP attacks.

3) This study demonstrated the efficiency of label-flipping DP attacks on EEG signal-based human emotion evaluation systems under complicated EEG signal features.

4) A deeper and explainable analysis of the consequences of the DP attacks on various ML models, comparing the results, quantifying the effects of the attack, and identifying each model's vulnerabilities with the help of XAI techniques.

5) This work analyzed different features' impacts on different Machine Learning models' human emotion prediction under DP attacks using XAI techniques.

The rest of this paper is organized as follows: Section II discusses the previous knowledge on various data poisoning attacks and the applications of EEG signals on human emotion evaluation. Section III then introduces the framework of the proposed label-flipping DP attacks to the Machine Learning models of EEG signal-based human emotion assessment systems. Section IV provides experimentation results and analysis in terms of different evaluations of conventional performance metrics and also the explanations of the DP attacks using XAI techniques. Section V concludes this paper and provides prospects and final remarks for future work.

## II. RELATED WORKS

There have been studies that concentrate on the topic of emotional reactions when seeing emotional changes. These studies offer crucial information for predicting people's responses based on changes in their brains.

In [5], the authors collected information from 17 people who were in a range of emotional states. The information was charted and categorized into four risk levels: low, normal, medium, and high. The five electrode Emotiv Insight EEG system is utilized since it is designed to be cost-effective. This study will classify anomalous EEG patterns that could indicate an insider threat and determine whether the employee is fit for duty.

In [12], a computer-centered diagnosis system was developed to diagnose Alzheimer's disease using EEG signals. To extract the features of EEG signals, the filtered signal was then divided into its frequency bands using the Discrete Wavelet Transform (DWT) approach. Many signal features, including logarithmic band power, standard deviation, variance, kurtosis, average energy, root mean square, and Norm, have been integrated into the DWT technique.

Signals from 32 people, including 16 women and 16 men, ranging in age from 19 to 37, are utilized in the most widely used DEAP dataset [13], which incorporates EEG. A total of 40 musical videos were played for the participants as stimuli during the experiment to gather data. Every video was created with the intention of stirring up strong feelings. There are 1,280 different raw data points in the dataset.

In this study [6], the authors proposed a novel method for emotion recognition that combines multichannel EEG analysis with a newly developed entropy called multivariate multiscale modified-distribution entropy (MM-mDistEn) with a model based on an artificial neural network (ANN) to outperform existing approaches. The suggested system outperformed previous approaches in tests using two distinct datasets.

For EEG features learning to perform emotional categorization, Chen *et al.* [14] presented a hierarchical bidirectional GRU model with an attention mechanism (H-ATT-BGRU). They demonstrated that, as compared to non-hierarchical models like CNN and LSTM, models that investigate hierarchical structures, such as H-AVE-BGRU and H-MAX-BGRU, perform better in classifying EEG features. They discovered that the H-AVE-BGRU model's classification accuracy is 8.4% and 1.9% higher than that of CNN and LSTM, respectively, for categorizing the valence feature and 8.1% and 2.5% for classifying the arousal feature baseline SVM model. The CNN model did not offer a significantly better outcome since its accuracy in categorizing the valence obtained 57.2%, which outperforms the BT model by 1% and its accuracy in classifying the arousal achieved 56.3%; this outperforms the SVM model by 1.8%. In classifying the valence and arousal, the LSTM model performed better than the CNN by 6.5% and 5.6%, respectively.

at would target the ML models used by the EEG analysis systems [15]. To the best of our knowledge, this research offers the first-ever and practically workable recommendation to use a narrow period pulse as a defense against poisoning attacks on EEG-based brain-computer interfaces (BCIs).

The most frequent method of producing this type of poisoning is through willful flipping of the data's labels [16]. Depending on the attacker's objectives, label flipping can be done randomly or deliberately. The former seeks to diminish the overall accuracy of all classes, while the latter does not focus on considerable accuracy reduction but rather on the misclassification of a specific class.

An effective label-flipping poisoning approach that compromises machine learning classifiers is suggested by Paudice *et al.* in their study [17]. Following an optimization formulation to optimize the target model's loss function, label-flipping operations are carried out. Due to the use of heuristic algorithms that allow label-flipping attacks to downscale the computing cost, this approach is regarded as computationally intractable.

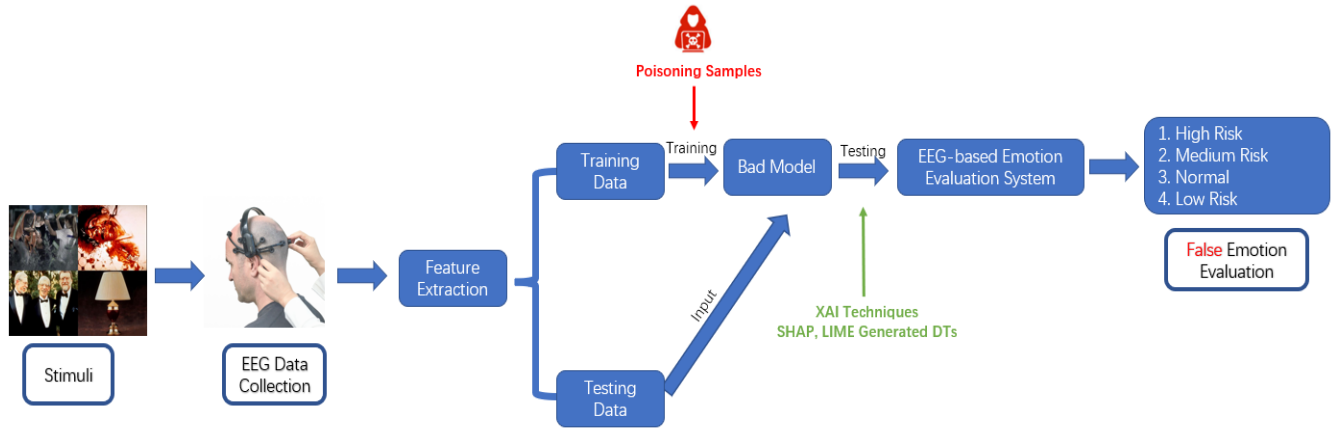

Fig. 1. The overview of the proposed explainable data poison attacks on EEG-based human emotion evaluation systems.

According to Xiao *et al.*[18], an SVM-based model has been successfully attacked utilizing label-flipping, which maximizes the possibility of classification errors. As a result, the classifier's overall accuracy has been significantly reduced. This method naturally indicates a significant computing overhead as a primary need, which could be a disadvantage.

On the other hand, from the defensors' perspective of label-flipping poisoning attacks, the k-Nearest-Neighbors defensive system [19] seeks to identify harmful material and mitigate its consequences; this defense is also known as label sanitization (LS). The defense of label sanitization (LS) is based on the decision boundary of SVM, which observes the distance of the poisoned samples and recommends that these samples be re-labeled.

As stated in [20], there is yet another method for anticipating outliers. Observe that the main priority in our aims is detection and mitigation activities. To prevent an attacker from tampering with the training set, the third sort of protection called data sanitization is intended to be used. As a result, we would pay greater attention to the data's quality before supplying it to the ML model of our choosing.

SVM is particularly prone to label-flipping assaults, which result in complete misclassification due to the computation of incorrect decision boundaries. SVM Resistance Enhancement [21] is designed to prevent these attacks. The suggested method takes a weighted SVM with KLID into account to anticipate the consequences of suspicious data points within the SVM decision border (K-LID-SVM). The K-LID approximation of the Local Intrinsic Dimensionality (LID) metric, which is linked to the outliers in data samples, is introduced in this paper. K-LID computation relies on the kernel distance involved in the LID calculation, allowing LID to be computed in high dimensional transformed spaces. obtaining the LID values in this manner, and as a result, identifying three distinct label dependant K-LID variations that can mitigate the effects of label flipping.

According to [22], poor generalization performance in Brain-Computer Interface (BCI) classification systems used in separate sessions can be caused by the non-stationarity of EEG signals. This paper concentrated on an experimental investigation of explanations generated by several XAI techniques using a machine learning system trained on a typical EEG dataset for emotion recognition. Results demonstrate that many pertinent elements discovered by XAI approaches are shared between sessions and can be used to create a system that can generalize more effectively.

In [23], the authors proposed an EEG signal analysis-based BCI system that can automatically recognize and decode voluntary eye blinks using Deep Learning. The primary goal of this study was to examine the explainability of the proposed CNN with the ultimate goal of determining which EEG signal segments are most crucial to the process of distinguishing between intentional and involuntary blinks. XAI methods were used to achieve this. In particular, the Local Interpretable Model Agnostic Explanation (LIME) and Gradient-weighted Class Activation Mapping (Grad-CAM) methods were applied. We were able to visually identify the most important EEG regions, particularly for the detection of voluntary and involuntary blinks, thanks to XAI.

In [24], A. Y. Al Hammadi *et al.* gathered one dataset to look into the potential applications of brainwave signals to industrial insider threat identification. The Emotiv Insight 5 channels device was used to connect the dataset. Data from 17 people who agreed to participate in the data collection are included in the dataset. The five Emotiv Insight electrodes, each with five power bands, will provide an EEG signal dataset. The total amount of data columns now equals 25 columns. There is also a timestamp column displayed. And this work will also deploy this dataset for investigation. The authors also implemented and compared Deep Learning techniques [25] and Machine Learning methods [1]. Furthermore, XAI techniques including Permutation Importance and SHAP values were explored in [5].

Although these previous works investigated the applications of using Artificial Intelligence techniques in EEG signal processing for human emotion evaluation, few researchers are concerned about the scenarios of DP attacks on EEG signal-based human emotion evaluation systems. These earlier results motivate further research into the utilization of human brainwave patterns to collect valuable data for identifying future DP attacks and accurately analyzing and explaining them using

Machine Learning algorithms and XAI techniques. Therefore, this paper proposed a framework of explainable DP attacks on human emotion evaluation systems based on EEG signals using XAI methods in Section III.

## III. RESEARCH METHODOLOGY

In this section, the methods utilized to build the framework of DP attacks on EEG signal-based human emotion assessment systems from the attackers' point of view are introduced, including the data processing of EEG signals, various ML models used to evaluate human emotions based on EEG signals, DP attacks based on label-flipping, and the XAI techniques deployed to explain the features of EEG signals. Figure. 1 provides a high-level view of the suggested architecture, while the subsections each provide a description of a particular component of the suggested attacking framework respectively.

### A. EEG Signal Data Collection and Processing

In accordance with the ethical guidelines established by the Khalifa University Compliance Committee, the EEG signal dataset was collected in a special facility at Khalifa university [24]. The dataset was gathered to investigate the possible uses of brainwave signals for identifying insider threats in the industrial setting. The dataset was gathered using a device called the Emotiv Insight 5 channels. This dataset includes information from 17 individuals who consented to participate in the data collection.

According to [26], most of n the observed EEG signals fall in the range of 1 – 100 Hz. And Delta, Theta, Alpha, Beta, and Gamma brainwave band frequency range from 0 Hz to 4 Hz, 4 Hz to 7 Hz, 8 Hz to 13 Hz, 14 Hz to about 30 Hz, and 31 Hz to 100 Hz respectively. Delta wave is seen normally in adults in slow wave sleep. It may manifest focally with diffuse lesions, diffuse subcortical lesions, metabolic encephalopathy hydrocephalus, or deep midline lesions, or it may manifest generally. Theta wave is seen normally in young children. In older children and adults, it may be observed in drowsiness or arousal, as well as in meditation. Theta wave excess for the age indicates aberrant activity. It can manifest as a diffuse disorder, metabolic encephalopathy, deep midline problems, focal disruption in focal subcortical lesions, or focal disturbance in some cases of hydrocephalus. Alpha wave was the "posterior basic rhythm", which was more pronounced on the dominant side and visible in the posterior regions of the skull on both sides. It appears when the eyes are closed and when one is relaxed, and it recedes when the eyes are opened or when one is working hard. Beta wave is most noticeable frontally and is typically distributed symmetrically on both sides. Beta activity is associated with motor behavior and is often reduced during vigorous motion.

The four risk categories—High-Risk, Medium-Risk, Low-Risk, and Normal—found in the risk matrix were used to classify each signal for a captured image, and each signal was then given the appropriate label. The data files' 26 features are made up of 25 inputs and 1 output. Each of the five electrodes on the EEG gadget records one of the five brainwave bands: Theta, Alpha, Low Beta, High Beta, and Gamma.

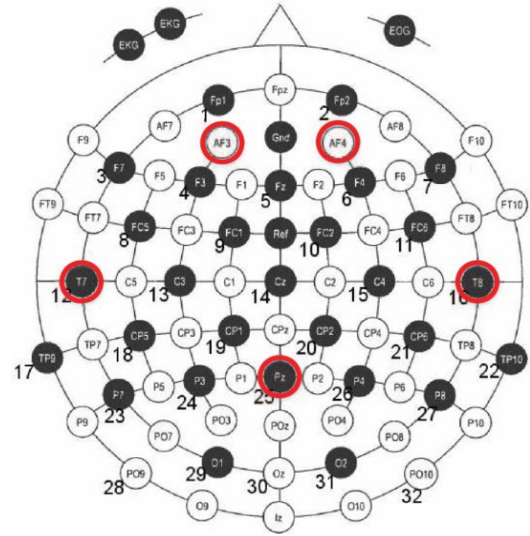

Fig. 2. Emotiv five channels positions.

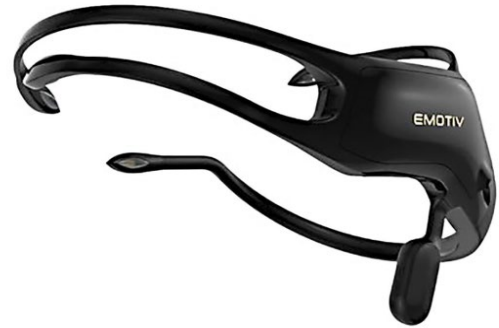

Fig. 3. Emotiv Insight device.

Figure 2 shows the electrodes in the Emotiv Insight device that record brainwaves related to cognitive and deceptive actions. And Figure 3 shows the Emotiv Insight device that was utilized to generate EEG signals. AF3 is connected to making decisions based on emotional cues, assuming others' intentions, and inferential reasoning. AF4 is responsible for decision-making involving incentives and conflicts, planning, and judgment. Pz deals with cognitive processes, whereas T7 and T8 deal with intentions. As a result, Table I displays the input data.

The four risk categories—High-Risk, Medium-Risk, Low-Risk, and Normal—found in the risk matrix were used to classify each signal for a captured image, and each signal was then given the appropriate label. The data files' 26 features are made up of 25 inputs and 1 output. Each of the five electrodes on the EEG gadget records one of the five brainwave bands: Theta, Alpha, Low Beta, High Beta, and Gamma.

Figure 2 shows the electrodes in the Emotiv Insight device that record brainwaves related to cognitive and deceptive actions. And Figure 3 shows the Emotiv Insight device that was utilized to generate EEG signals. AF3 is connected to making decisions based on emotional cues, assuming others' intentions, and inferential reasoning. AF4 is responsible for decision-making involving incentives and conflicts, planning, and judgment. Pz deals with cognitive processes, whereas T7 and T8 deal with intentions. As a result, Table I displays the input data.

*B. DP Attacks on Machine Learning Models*

Six different ML models—Random Forest, Extra Trees, AdaBoost, MLP, XGBoost, and KNN—are intruded on by DP attacks. These six ML classifiers were applied for the classification tasks of risk assessment based on the obtained EEG signal data without DP attacks initially. The dataset has been split into 80% for training the ML models, and 20% for testing.

Label-flipping technique has been offered as the DP attack on the training sets of EEG signal data, as we covered in the preceding part. Given the attacker's capacity to inject poison samples in vast amounts, it stands to reason that the label-flipping attacks with the fewest harmful samples are the most effective. A key factor to consider while evaluating the training set is the ratio of poisoned samples to all samples. Additionally, from the attackers' point of view, it is beneficial to assess the various resilience capacities of various ML models used in the classification based on the EEG signal data. The training set was therefore subjected to various poisoning rates, including 5%, 25%, 50%, and 75%, and the results were compared to the models' initial performance (0% poisoning).

In addition, two distinct Label Flipping scenarios are described in this paper because there are four labels in the risk assessment system task based on the EEG signal data. Other categories like Medium-Risk, Normal, and Low-Risk would change to High-Risk in the first scenario. This is done to see how changing other risk categories to a High-Risk level may affect the human emotion assessment system. The second scenario, on the other hand, would see Low-Risk turn into Normal, Normal turn into Medium-Risk, Medium-Risk turn into High-Risk, and High-Risk turn into Low Risk. In a similar manner, the second scenario is used to observe the impact on the human emotion evaluation system based on EEG signals after altering danger levels to the other immediate risk levels.

*C. Explainability of the Model*

The creation of an explainable framework that will look into the essential features of the suggested DP attacks is one of the objectives of this research. Explainability assures that the obtained DP attacks emerge from explainable conditions rather than a black-box operation and not just improve understanding of the machine learning models.

This paper will use XAI techniques including Shapley Additive Explanation (SHAP), Local Interpretable Model-agnostic Explanations (LIME), and Generated Decision Trees to describe the proposed DP attacks framework.

TABLE I
List of All Inputs from Emotiv Electrodes.

| Electrode | Input features |
|---|---|
| AF3 | AF3_THETA, AF_ALPHA, AF3_LOW_BETA, AF3_HIGH_BETA, AF3_GAMMA |
| T7 | T7_THETA, T7_ALPHA, T7_LOW_BETA, T7_HIGH_BETA, T7_GAMMA |
| Pz | Pz_THETA, Pz_ALPHA, Pz_LOW_BETA, Pz_HIGH_BETA, Pz_GAMMA |
| T8 | T8_THETA, T8_ALPHA, T8_LOW_BETA, T8_HIGH_BETA, T8_GAMMA |
| AF4 | AF4_THETA, AF4_ALPHA, AF4_LOW_BETA, AF4_HIGH_BETA, AF4_GAMMA |

SHAP was introduced as a model-neutral approach to discussing machine learning models in 2017 [27]. Shapley values are determined by comparing the team's performance with and without each individual player in a team game to determine their contribution. By evaluating the difference between the model performance with and without the feature, this method in machine learning determines the influence of each feature. This clarifies the extent to which each feature contributes positively or negatively to the forecast. As stated in [28], SHAP values are thought to be a superior technique of explanation to feature importance. The process of assigning a score to each input characteristic for a certain model is known as feature significance. The relevance of each characteristic is indicated by its score. Only machine learning algorithms can calculate feature importance, which is determined by Gini importance using node impurity.

Local Interpretable Model-agnostic Explanations (LIME) were suggested by Marco *et al.* in [29]. The primary goal of the LIME approach is to find an interpretable model over the interpretable representation that is both locally true to the classifier and understandable to humans. For a classifier (complex model), you want to use an interpretable model (simple model such as linear programming) with interpretable features for adaptation, and this interpretable model is then locally close to the effect of the complex model in terms of performance.

By deploying the generated decision tree models, this paper also explored the XAI idea to improve trust management. This study divided the choice into numerous little subchoices for human emotion evaluation based on EEG data, using straightforward decision tree algorithms that are simple to read and even mimic a human approach to decision-making. To test this strategy, we extracted rules from a dataset that was being used.

TABLE II
Metrics Comparison among ML Models for the First Category of the DP Attack

| ML model | Poison rate [%] | Accuracy [%] | Recall [%] | Precision [%] | F1-Score [%] | Log loss |
|---|---|---|---|---|---|---|
| Ada Boost | 0 | 99.68 | 99.67 | 99.67 | 99.66 | 0.017 |
| | 5 | 96.77 | 96.85 | 96.96 | 96.76 | 0.097 |
| | 25 | 73.87 | 74.67 | 61.68 | 65.99 | 7.523 |
| | 50 | 50.32 | 50.00 | 33.19 | 37.34 | 15.282 |
| | 75 | 24.19 | 25.00 | 6.05 | 9.74 | |
| Random Forest | 0 | 90.97 | 91.03 | 91.94 | 91.04 | 0.627 |
| | 5 | 86.77 | 86.94 | 88.49 | 86.87 | 0.697 |
| | 25 | 60.97 | 61.93 | 55.66 | 55.33 | 3.930 |
| | 50 | 34.51 | 34.88 | 31.74 | 24.78 | 5.223 |
| | 75 | 24.19 | 25.00 | 6.05 | 9.74 | |
| MLP | 0 | 76.13 | 75.97 | 76.24 | 75.67 | 0.561 |
| | 5 | 43.55 | 43.66 | 47.10 | 43.67 | 1.928 |
| | 25 | 42.26 | 43.38 | 39.99 | 35.99 | 7.314 |
| | 50 | 39.68 | 39.62 | 26.73 | 29.09 | 10.238 |
| | 75 | 24.19 | 25.00 | 6.05 | 9.74 | |
| KNN | 0 | 80.65 | 80.57 | 80.48 | 80.43 | 2.138 |
| | 5 | 78.71 | 78.69 | 78.70 | 78.63 | 2.409 |
| | 25 | 57.42 | 58.11 | 48.35 | 51.31 | 10.560 |
| | 50 | 41.61 | 41.52 | 27.61 | 30.48 | 17.719 |
| | 75 | 24.19 | 25.00 | 6.05 | 9.74 | |
| ExtraTree | 0 | 99.35 | 99.33 | 99.34 | 99.33 | 0.217 |
| | 5 | 95.48 | 95.58 | 95.81 | 95.45 | 0.286 |
| | 25 | 73.55 | 74.33 | 61.27 | 65.68 | 8.591 |
| | 50 | 49.68 | 49.38 | 33.12 | 36.94 | 16.553 |
| | 75 | 24.19 | 25.00 | 6.05 | 9.74 | |
| XG Boost | 0 | 85.48 | 85.54 | 86.31 | 85.46 | 0.498 |
| | 5 | 84.19 | 84.39 | 85.12 | 84.24 | 0.577 |
| | 25 | 60.65 | 61.53 | 52.53 | 54.54 | 2.074 |
| | 50 | 37.10 | 37.32 | 31.31 | 27.45 | 3.554 |
| | 75 | 24.19 | 25.00 | 6.05 | 9.74 | |

TABLE III
Metrics Comparison among ML Models for the Second Category of the DP Attack

| ML model | Poison rate [%] | Accuracy [%] | Recall [%] | Precision [%] | F1-Score [%] | Log loss |
|---|---|---|---|---|---|---|
| Ada Boost | 0 | 99.68 | 99.67 | 99.67 | 99.66 | 0.017 |
| | 5 | 96.77 | 96.79 | 96.79 | 96.74 | 0.118 |
| | 25 | 76.45 | 76.46 | 76.67 | 76.33 | 0.722 |
| | 50 | 57.74 | 58.25 | 63.59 | 56.61 | 1.799 |
| | 75 | 21.61 | 21.76 | 21.53 | 21.53 | 4.136 |
| Random Forest | 0 | 90.97 | 91.03 | 91.94 | 91.04 | 0.627 |
| | 5 | 88.71 | 88.85 | 89.87 | 88.79 | 0.659 |
| | 25 | 80.00 | 80.17 | 81.19 | 80.05 | 0.858 |
| | 50 | 50.00 | 50.56 | 57.41 | 48.02 | 0.995 |
| | 75 | 14.52 | 14.52 | 14.25 | 14.32 | 1.490 |
| MLP | 0 | 76.13 | 75.97 | 76.24 | 75.67 | 0.561 |
| | 5 | 43.55 | 43.92 | 46.16 | 43.42 | 1.250 |
| | 25 | 39.68 | 39.41 | 39.42 | 39.15 | 1.772 |
| | 50 | 34.84 | 35.41 | 39.80 | 30.73 | 2.678 |
| | 75 | 16.77 | 16.75 | 16.43 | 15.99 | 3.444 |
| KNN | 0 | 80.65 | 80.57 | 80.48 | 80.43 | 2.138 |
| | 5 | 79.35 | 79.31 | 79.39 | 79.18 | 2.411 |
| | 25 | 68.39 | 68.37 | 69.47 | 68.20 | 3.577 |
| | 50 | 50.00 | 50.38 | 49.92 | 48.56 | 6.625 |
| | 75 | 18.06 | 18.02 | 17.76 | 17.75 | 16.101 |
| Extra Tree | 0 | 99.35 | 99.33 | 99.34 | 99.33 | 0.217 |
| | 5 | 95.48 | 95.49 | 95.47 | 95.44 | 0.287 |
| | 25 | 78.39 | 78.25 | 78.73 | 78.03 | 0.606 |
| | 50 | 60.32 | 60.64 | 64.19 | 59.62 | 1.107 |
| | 75 | 20.97 | 21.07 | 20.95 | 20.93 | 2.342 |
| XG Boost | 0 | 85.48 | 85.54 | 86.31 | 85.46 | 0.498 |
| | 5 | 84.10 | 84.09 | 84.55 | 84.02 | 0.534 |
| | 25 | 75.16 | 75.23 | 76.34 | 75.21 | 0.803 |
| | 50 | 54.19 | 54.71 | 60.89 | 53.52 | 0.977 |
| | 75 | 14.52 | 14.53 | 13.45 | 13.94 | 1.591 |

## IV. EXPERIMENTAL RESULTS AND ANALYSIS

### A. Performance Metrics

A Machine Learning-based classifier's four primary performance indicators are as follows:

1) The proportion of test instances with true and expected values of 1 divided by the proportion of test instances having a true value of 1 is known as the True Positive (TP) rate.

2) The number of test instances with true and anticipated values of 0 divided by the total number of test instances having a true value of 0 is known as the True Negative (TN).

3) The ratio of the number of test instances with a true value of 0 and a predicted value of 1 to the number of instances with a true value of 0 is known as the False Positive (FP).

4) The number of test instances with a true value of 1 and a predicted value of 0 divided by the number of test instances with a true value of 1 is known as the False Negative (FN).

These four measurements work together to create the confusion matrix. The confusion matrix is used to assess the effectiveness of the filtering models, and FP, FN, TP, and TN are defined as follows. The following equations represent the statistical measures, which include Accuracy, Recall, Precision, and F1-score:

$$Accuracy = \frac{TP + TN}{TP + TN + FP + FN} \quad (1)$$

$$Recall = \frac{TP}{TP + FN} \quad (2)$$

$$Precision = \frac{TP}{TP + FP} \quad (3)$$

$$F1 - Score = \frac{2 \times (Precison \times Recall)}{Precison + Recall} \quad (4)$$

Other than the performance metrics illustrated, log loss is also utilized in this work to measure the loss of classification abilities of the Machine Learning classifiers under the different levels of DP attacks. Log loss indicates how closely the forecast probability matches the associated real or true value. The higher the log-loss number, the more the predicted probability deviates from the actual value. And the log loss function $F$ in terms of the logarithmic loss function per label $F_i$ is defined as:

$$F = -\frac{1}{N}\sum_i^N \sum_j^M \left( y_{ij} \cdot Ln(p_{ij}) \right) = \sum_J^M \left( -\frac{1}{N}\sum_i^N y_{ij} \cdot Ln(p_{ij}) \right) = \sum_j^M F_i \quad (5)$$

Where N is the number of instances, M is the number of different labels, $y_{ij}$ is the binary variable with the expected labels and $p_{ij}$ is the classification probability output by the classifier for the i-instance and the j-label.

### B. DP Attack Results

This section presents the results of the proposed DP attack scenarios against the EEG-based risk assessment system.

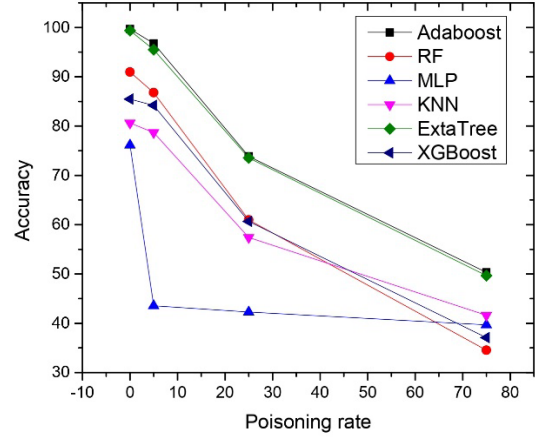

Fig. 4. Accuracy comparison among ML models under the first category of DP attack.

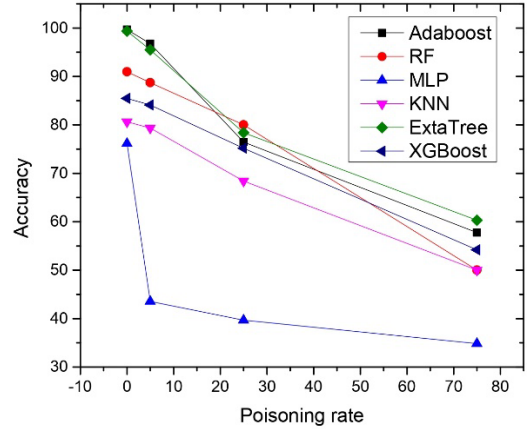

Fig. 5. Accuracy comparison among ML models under the second category of DP attack.

Table II and Table III highlight the metrics comparison for the proposed DP attack scenarios under different poisoning rates among six different ML models—Random Forest, Extra Trees, AdaBoost, MLP, XGBoost, and KNN. These metrics include the classification Accuracy of the ML models, Recall, Precision, F1 score, and Log loss. As presented in the tables, although, the six ML models show different resilience to the DP attacks, the attacks were able to degrade the performance of the ML models. For example, in terms of classification accuracy, AdaBoost archived an accuracy of 99.68% with a 0% poisoning rate. Consequently, the accuracy of AdaBoost degraded to 24.19% with a 75% poisoning rate. A similar trend is observed among all other ML models.

Figures 4 and 5 present the accuracy metric comparison for the proposed DP attacks scenarios under different poisoning rates among six different ML models. According to the accuracy metric shown in the figures, AdaBoost and ExtraTree models archived relatively higher accuracy compared to the other ML models. Figures 6 and 7 present the confusion

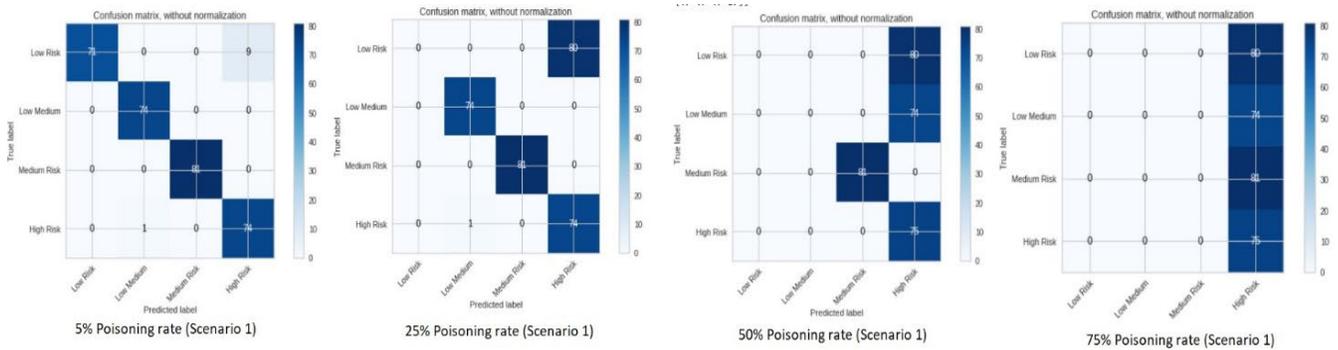

Fig. 6. Confusion matrix for Random Forest ML model under the first category of DP attack.

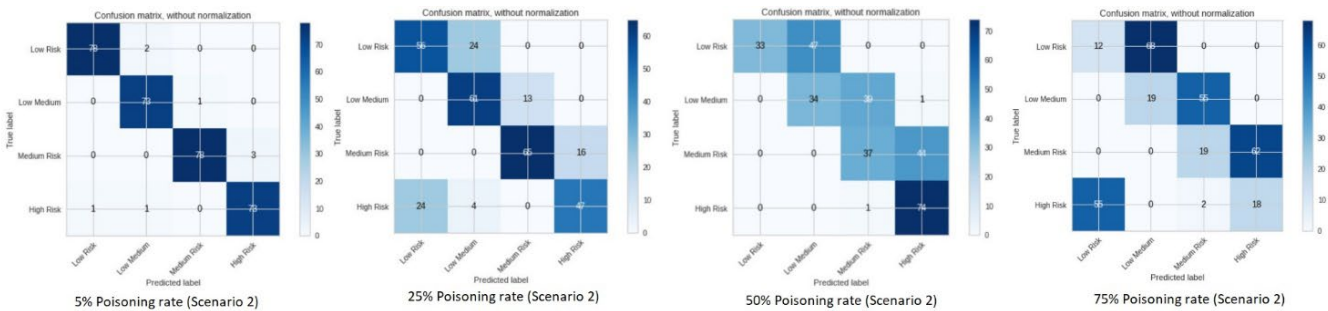

Fig. 7. Confusion matrix for Random Forest ML model under the second category of DP attack.

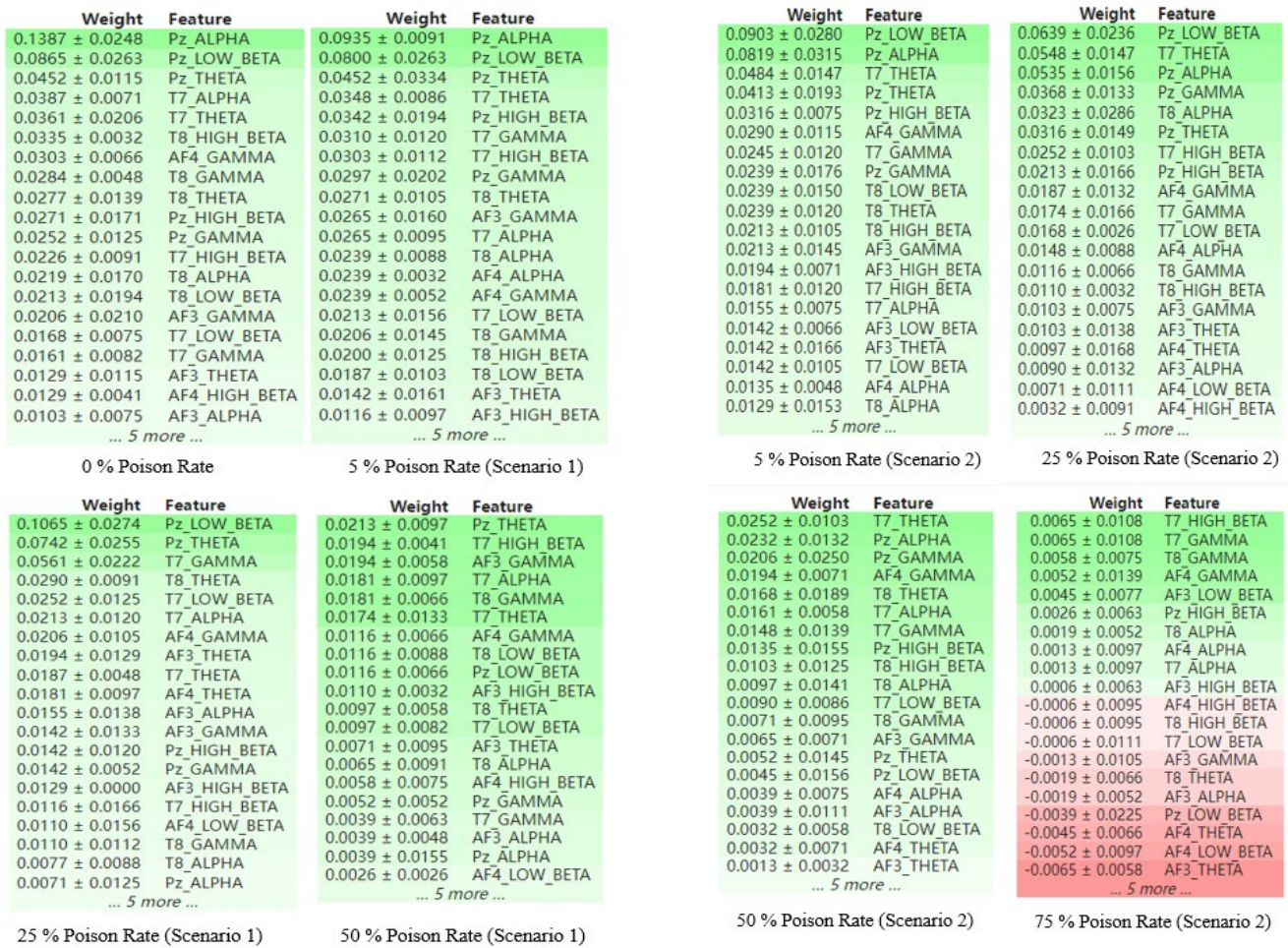

Fig. 8. Permutation importance of Random Forest under DP attack Scenario 1.

Fig. 9. Permutation importance of Random Forest under DP attack Scenario 2.

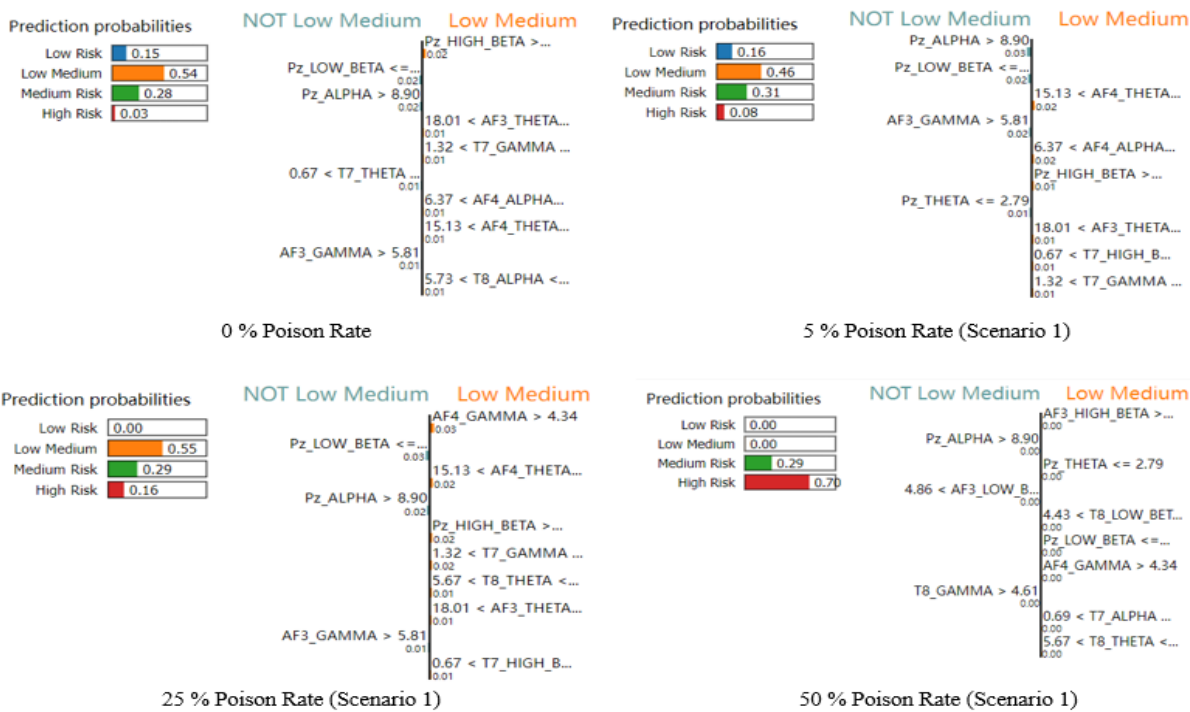

Fig. 10. Prediction probabilities and feature contribution based on LIME of Random Forest under DP attack Scenario 1.

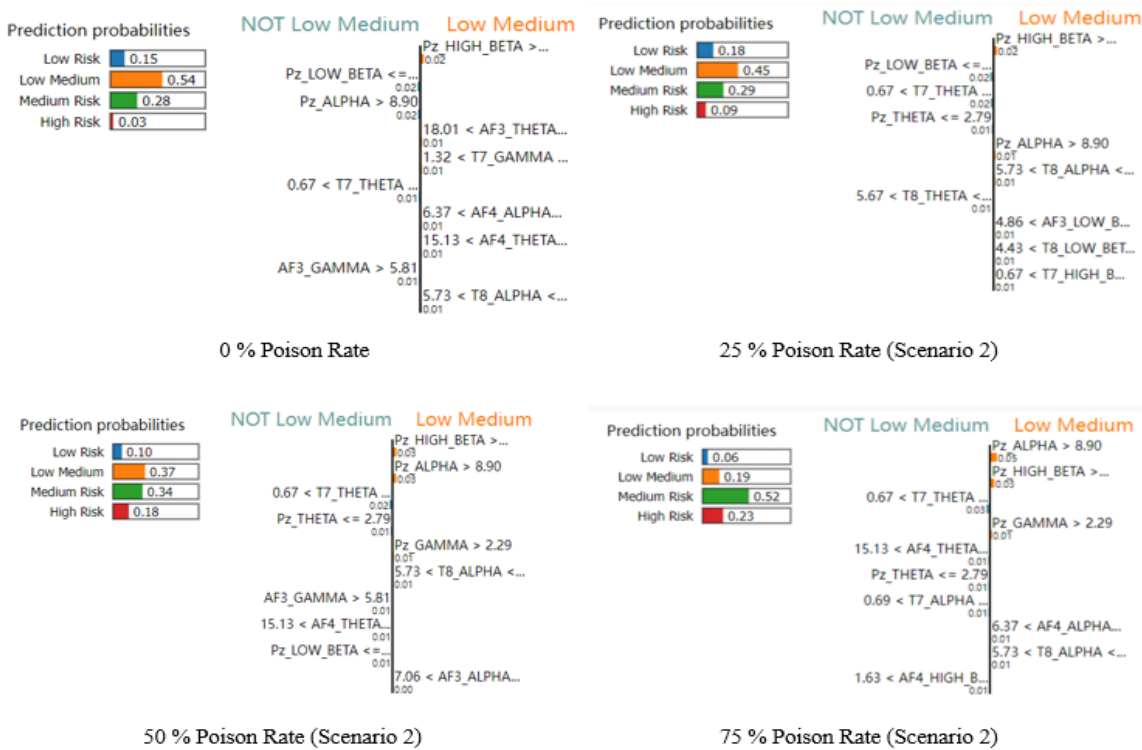

Fig. 11. Prediction probabilities and feature contribution based on LIME of Random Forest under DP attack Scenario 2.

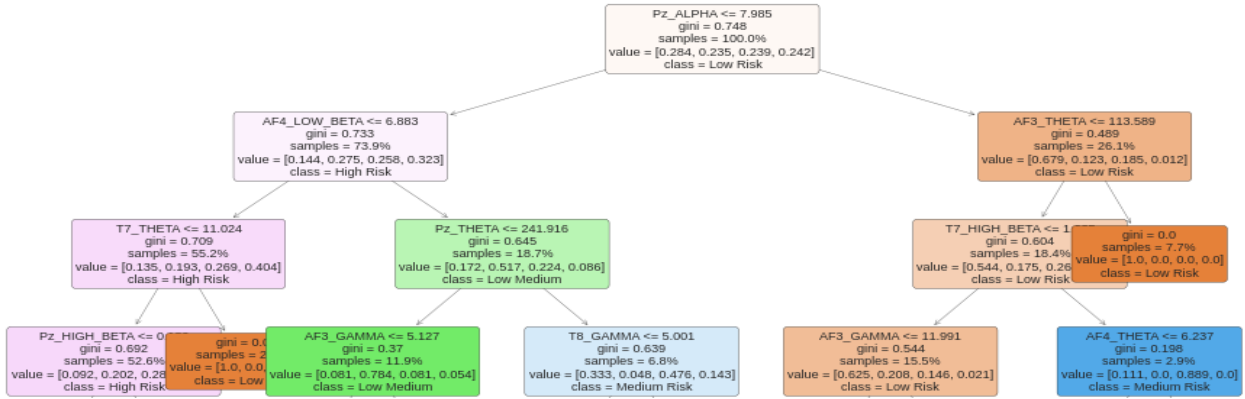

Fig. 12. Explainable results of Random Forest classifier under DP attack Scenario 1 using Generated Decision Trees.

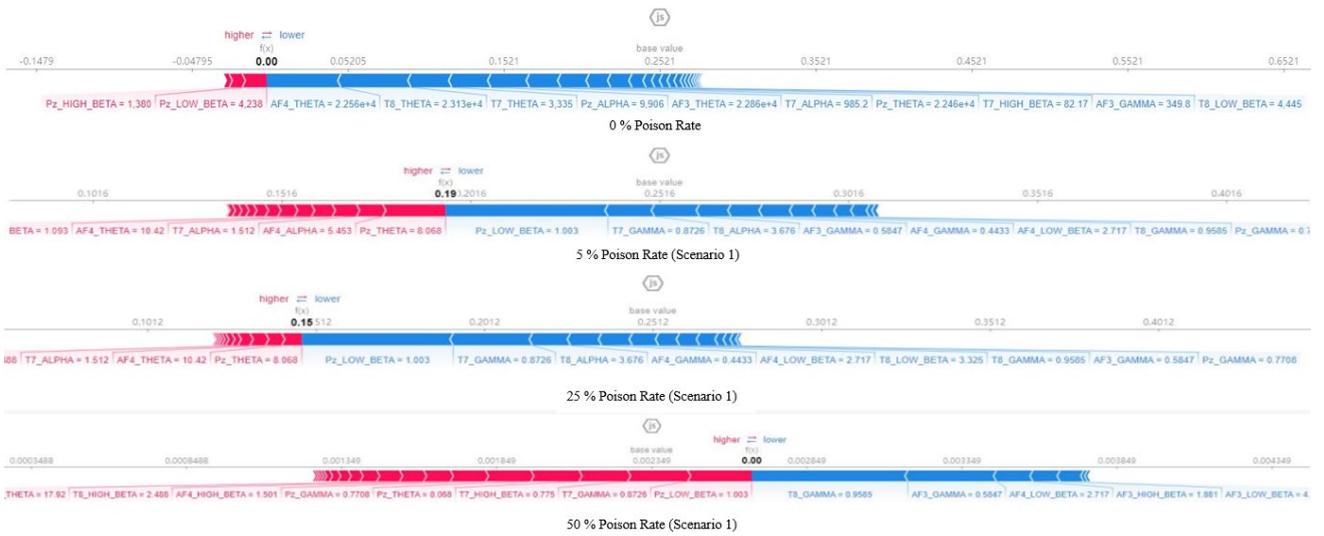

Fig. 13. Explainable results of Random Forest classifier under DP attack Scenario 1 using SHAP Force Plot.

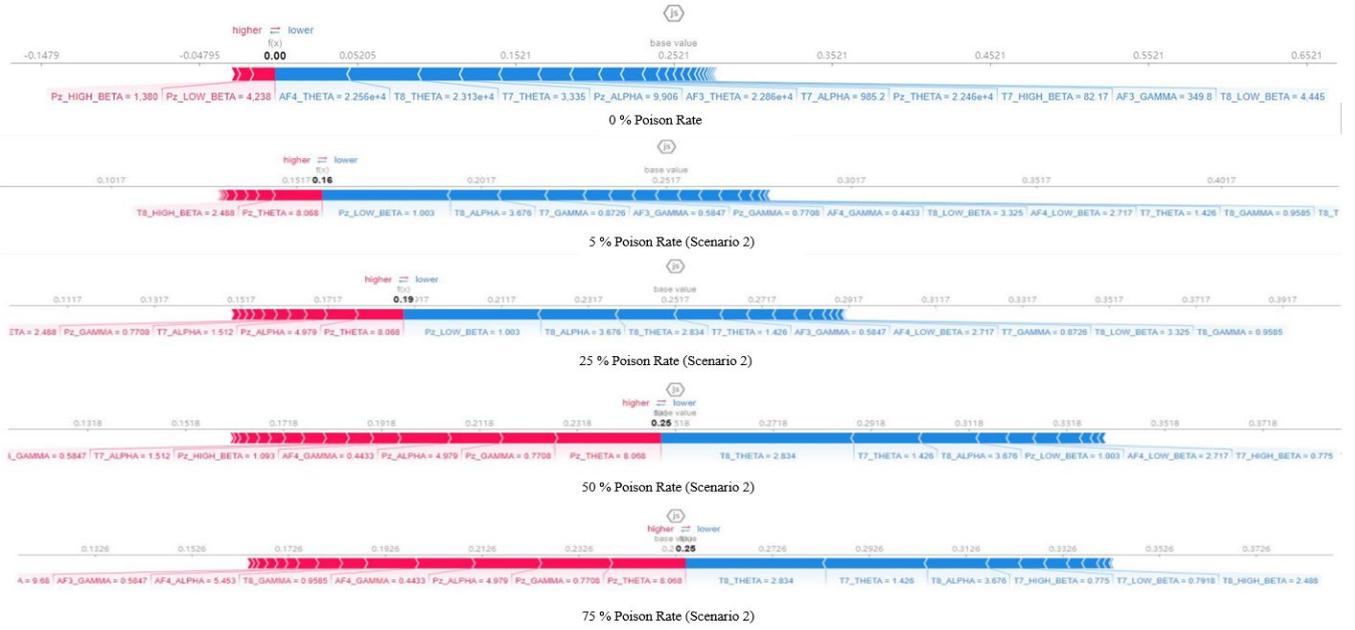

Fig. 14. Explainable results of Random Forest classifier under DP attack Scenario 2 using SHAP Force Plot.

matrix for the Random Forest ML model under different poisoning rates for the proposed DP attack scenarios. The confusion matrix illustrates an increase in misclassification as the poisoning rates increase from five percent (5%) to seventy-five percent (75%). In the next section, we will present the explanation results of the proposed two DP attack scenarios.

### C. Explanation Results

This section presents the explanation results of the proposed two DP attack scenarios against the human emotion evaluation systems based on EEG signals. To explain the employed Machine Learning models' behaviors and the key features that ultimately determined the classification results of the algorithm, this paper deployed several XAI techniques including SHAP, LIME, Generated Trees, and the Permutation Importance to explain the effects of different features under the proposed two DP attacks.

According to the Permutation Importance of Random Forest under DP attack Scenario 1 illustrated in Figure 8, the permutation importance shows that Pz_ALPHA, Pz_LOW_BETA, and Pz_THETA are the most significant features for the classification of the Random Forest algorithm whereas other features have lower but also positive effects on the decisions of the Random Forest classifier. Under a 5% scenario 1 poison rate, the feature importance ranking has few changes whereas under a 25% scenario 1 poison rate and a 50% scenario 1 poison rate, Pz_LOW_BETA, Pz_THETA, T7_GAMMA and Pz_THETA, T7_HIGH_BETA, AF3_GAMMA are the most significant features respectively. However, other features still contribute positively to the classification decision under all circumstances. For the 75% scenario 1 poison rate, Permutation Importance could not work as all samples are classified into one class in this situation.

According to the Permutation Importance of Random Forest under DP attack Scenario 2 illustrated in Figure 9, for the classification of the Random Forest classifier, under a 5% scenario 2 poison rate, a 25% scenario 2 poison rate, and a 50% scenario 2 poison rate, Pz_LOW_BETA, Pz_ALPHA, T7_THETA, Pz_LOW_BETA, T7_ THETA, Pz_ALPHA, and T7_THETA, Pz_ALPHA, Pz_GAMMA are the most significant 3 features respectively. Moreover, other features still contribute positively to the classification decision under all circumstances. However, for the 75% scenario 22 poison rate, Permutation Importance shows that T7_HIGH_BETA, T7_GAMMA, and T8_GAMMA contribute most to the classification results. On the other hand, half of the features including AF3_THETA, AF4_LOW_BETA, and AF4_THETA make negative contributions to the final classification results.

More explainable results explained by the Permutation Importance of other Machine Learning algorithms are shown in the appendix.

As shown in Figure 10 and Figure 11, the prediction probabilities and feature contribution based on LIME of Random Forest under DP attack Scenario 1 and 2 illustrate slight differences. In scenario 1, the Random Forest classifier made false predictions in the 50% poison rate whereas the Random Forest classifier made false predictions in the 75% poison rate in scenario 2. More explainable results explained by the LIME of other Machine Learning algorithms are shown in the appendix.

Moreover, this paper used Explainable AI to create Decision Trees (Figure 12) using the IBM SPSS Modeler tool and determine the features that should be prioritized for the classifier. This allowed us to simulate the steps that a possible attacker would take. More explainable results explained by the Generated Decision Trees of other Machine Learning algorithms are shown in the appendix.

This paper also created the SHAP value to help us achieve the goal of integrating Explainable AI. The SHAP value calculates the strength and polarity (positive or negative) of a feature's influence on a prediction. Model semantics can be clearly expressed by SHAP values. SHAP power sites offer a thorough perspective of the model's internal operation, i.e., they demonstrate the model's decision-making process. Figure 13 and Figure 14 depict the impact of a specific record. The model's lower decision in the chosen record is the forecast from the first phase in this situation. The framework outlines the elements that support a given choice. By using this data, the decision-making process of the model can be contrasted with that in the real world and ensured that the model bases its decisions on pertinent information.

## V. CONCLUSION

This study proposed two label-flipping attacks on human emotion evaluation systems based on EEG brainwave signals. Two scenarios of the label-flipping attack were analyzed and evaluated to drastically decrease the overall accuracy and misclassification rate of six different multi-label Machine Learning classifiers including MLP, AdaBoost, Random Forest, KNN, XGBoost, and Extra Trees. On the other hand, the influence of training data poisoning attacks, such as the proposed label-flipping attacks varies on different Machine Learning classifiers and different EEG signal features. The use of AdaBoost, Random Forest, MLP, KNN, Extra Trees, and XGBoost classifiers achieved accuracies of 96.77 %, 86.77%, 43.55%, 78.71%, 95.48%, and 84.19% respectively under 5% poisoning rate. The accuracies of these classifiers decreased to 73.87%, 60.97%, 42.26%, 57.42%, 73.55%, and 60.65% under 25% poisoning rate and 50.32%, 34.51%, 39.68%, 41.61%, 49.68%, and 37.10% under 50% poisoning rate.

To better analyze the impacts of the proposed two label-flipping attacks on different features of EEG brainwave signals and the emotion classification results, this study utilized several XAI techniques including SHAP, LIME, Generated Decision Trees, and Permutation Importance. The results show that the AdaBoost classifier performs best in terms of classification and poisoning resilience. Besides, features of the Pz electrode contribute more to the classification results of emotion assessment according to the XAI technique, Permutation Importance. Moreover, the scenario 1 DP attack can damage the classifier with a lower poisoning rate compared with the scenario 2 DP attack according to the XAI technique, LIME. More details about the explainable results of the label-flipping attacks on human emotion evaluation systems based on EEG brainwave signals using XAI techniques are available in the appendix.

Furthermore, the main future work goal of this research is to increase the robustness of Machine Learning models against tampered training data to be used during re-training with the method of XAI technique. Besides, DP attacks based on label-flipping during the testing phase will be investigated in future work. The resilience of Deep Learning methods on the DP attacks will be evaluated and compared in future work as well. Because this paper concluded that not all features contribute to the human emotion classification prediction, EEG signal generating with fewer features but similar accuracy will be explored in future work.

## VI. APPENDIX

Codes and detailed explainable results with XAI are publicly available on GitHub:

https://github.com/qiuyuezhibo/XAIEEG

## REFERENCES

[1] A. Y. Al Hammadi et al., "Novel EEG Sensor-Based Risk Framework for the Detection of Insider Threats in Safety Critical Industrial Infrastructure," IEEE Access, vol. 8, pp. 206222–206234, 2020, doi: 10.1109/ACCESS.2020.3037979.

[2] B. A. Erol, A. Majumdar, P. Benavidez, P. Rad, K.-K. R. Choo, and M. Jamshidi, "Toward Artificial Emotional Intelligence for Cooperative Social Human–Machine Interaction," IEEE Trans. Comput. Soc. Syst., vol. 7, no. 1, pp. 234–246, Feb. 2020, doi: 10.1109/TCSS.2019.2922593.

[3] K. Mohan, A. Seal, O. Krejcar, and A. Yazidi, "Facial Expression Recognition Using Local Gravitational Force Descriptor-Based Deep Convolution Neural Networks," IEEE Trans. Instrum. Meas., vol. 70, pp. 1–12, 2021, doi: 10.1109/TIM.2020.3031835.

[4] M. Karnati, A. Seal, A. Yazidi, and O. Krejcar, "LieNet: A Deep Convolution Neural Network Framework for Detecting Deception," IEEE Trans. Cogn. Dev. Syst., vol. 14, no. 3, pp. 971–984, Sep. 2022, doi: 10.1109/TCDS.2021.3086011.

[5] A. Y. Al Hammadi et al., "Explainable artificial intelligence to evaluate industrial internal security using EEG signals in IoT framework," Ad Hoc Netw., vol. 123, p. 102641, Dec. 2021, doi: 10.1016/j.adhoc.2021.102641.

[6] S. T. Aung et al., "Entropy-Based Emotion Recognition from Multichannel EEG Signals Using Artificial Neural Network," Comput. Intell. Neurosci., vol. 2022, p. e6000989, Oct. 2022, doi: 10.1155/2022/6000989.

[7] M.-P. Hosseini, A. Hosseini, and K. Ahi, "A Review on Machine Learning for EEG Signal Processing in Bioengineering," IEEE Rev. Biomed. Eng., vol. 14, pp. 204–218, 2021, doi: 10.1109/RBME.2020.2969915.

[8] M. A. Ramirez et al., "New data poison attacks on machine learning classifiers for mobile exfiltration." arXiv, Oct. 20, 2022. doi: 10.48550/arXiv.2210.11592.

[9] N. Pitropakis, E. Panaousis, T. Giannetsos, E. Anastasiadis, and G. Loukas, "A taxonomy and survey of attacks against machine learning," Comput. Sci. Rev., vol. 34, p. 100199, Nov. 2019, doi: 10.1016/j.cosrev.2019.100199.

[10] A. Paudice, L. Muñoz-González, A. Gyorgy, and E. C. Lupu, "Detection of Adversarial Training Examples in Poisoning Attacks through Anomaly Detection." arXiv, Feb. 08, 2018. doi: 10.48550/arXiv.1802.03041.

[11] Z. Zhang, H. A. Hamadi, E. Damiani, C. Y. Yeun, and F. Taher, "Explainable Artificial Intelligence Applications in Cyber Security: State-of-the-Art in Research," IEEE Access, vol. 10, pp. 93104–93139, 2022, doi: 10.1109/ACCESS.2022.3204051.

[12] K. AlSharabi, Y. Bin Salamah, A. M. Abdurraqeeb, M. Aljalal, and F. A. Alturki, "EEG Signal Processing for Alzheimer's Disorders Using Discrete Wavelet Transform and Machine Learning Approaches," IEEE Access, vol. 10, pp. 89781–89797, 2022, doi: 10.1109/ACCESS.2022.3198988.

[13] S. Koelstra et al., "DEAP: A Database for Emotion Analysis ;Using Physiological Signals," IEEE Trans. Affect. Comput., vol. 3, no. 1, pp. 18–31, Jan. 2012, doi: 10.1109/T-AFFC.2011.15.

[14] J. X. Chen, D. M. Jiang, and Y. N. Zhang, "A Hierarchical Bidirectional GRU Model With Attention for EEG-Based Emotion Classification," IEEE Access, vol. 7, pp. 118530–118540, 2019, doi: 10.1109/ACCESS.2019.2936817.

[15] L. Meng et al., "EEG-Based Brain-Computer Interfaces Are Vulnerable to Backdoor Attacks." arXiv, Jan. 02, 2021. doi: 10.48550/arXiv.2011.00101.

[16] X. Liu, H. Li, G. Xu, Z. Chen, X. Huang, and R. Lu, "Privacy-Enhanced Federated Learning Against Poisoning Adversaries," IEEE Trans. Inf. Forensics Secur., vol. 16, pp. 4574–4588, 2021, doi: 10.1109/TIFS.2021.3108434.

[17] A. Paudice, L. Muñoz-González, and E. C. Lupu, "Label Sanitization Against Label Flipping Poisoning Attacks," in ECML PKDD 2018 Workshops, Cham, 2019, pp. 5–15. doi: 10.1007/978-3-030-13453-2_1.

[18] B. Biggio, B. Nelson, and P. Laskov, "Poisoning Attacks against Support Vector Machines." arXiv, Mar. 25, 2013. doi: 10.48550/arXiv.1206.6389.

[19] B. Biggio, B. Nelson, and P. Laskov, "Support Vector Machines Under Adversarial Label Noise," in Proceedings of the Asian Conference on Machine Learning, Nov. 2011, pp. 97–112. Accessed: Dec. 16, 2022. [Online]. Available: https://proceedings.mlr.press/v20/biggio11.html

[20] "Mitigating Poisoning Attacks on Machine Learning Models | Proceedings of the 10th ACM Workshop on Artificial Intelligence and Security." https://dl.acm.org/doi/abs/10.1145/3128572.3140450?casa_token=hDUL68u9M5EAAAAA:sWwryFXB-dolE3qusCIUoUKE9C3PP7zmbRTH4u0HHqb4p3txtLFVJKfj7-2WjVTJq1NNl_SARoP3 (accessed Dec. 16, 2022).

[21] S. Chen et al., "Automated poisoning attacks and defenses in malware detection systems: An adversarial machine learning approach," Comput. Secur., vol. 73, pp. 326–344, Mar. 2018, doi: 10.1016/j.cose.2017.11.007.

[22] A. Apicella, F. Isgrò, A. Pollastro, and R. Prevete, "Toward the application of XAI methods in EEG-based systems." arXiv, Nov. 05, 2022. doi: 10.48550/arXiv.2210.06554.

[23] M. L. Giudice et al., "Visual Explanations of Deep Convolutional Neural Network for eye blinks detection in EEG-based BCI applications," in 2022 International Joint Conference on Neural Networks (IJCNN), Jul. 2022, pp. 01–08. doi: 10.1109/IJCNN55064.2022.9892567.

[24] A. A. AlHammadi, "EEG Brainwave Dataset." IEEE, Feb. 25, 2021. Accessed: Nov. 18, 2022. [Online]. Available: https://ieee-dataport.org/documents/eeg-brainwave-dataset

[25] A. Y. Al Hammadi, C. Yeob Yeun, and E. Damiani, "Novel EEG Risk Framework to Identify Insider Threats in National Critical Infrastructure Using Deep Learning Techniques," in 2020 IEEE International Conference on Services Computing (SCC), Nov. 2020, pp. 469–471. doi: 10.1109/SCC49832.2020.00071.

[26] B. R. Cahn and J. Polich, "Meditation states and traits: EEG, ERP, and neuroimaging studies," Psychol. Bull., vol. 132, no. 2, pp. 180–211, Mar. 2006, doi: 10.1037/0033-2909.132.2.180.

[27] S. M. Lundberg and S.-I. Lee, "A Unified Approach to Interpreting Model Predictions," in Advances in Neural Information Processing Systems, 2017, vol. 30. Accessed: Jul. 09, 2022. [Online]. Available: https://proceedings.neurips.cc/paper/2017/hash/8a20a8621978632d76c43dfd28b67767-Abstract.html

[28] W. E. Marcílio and D. M. Eler, "From explanations to feature selection: assessing SHAP values as feature selection mechanism," in 2020 33rd SIBGRAPI Conference on Graphics, Patterns and Images (SIBGRAPI), Nov. 2020, pp. 340–347. doi: 10.1109/SIBGRAPI51738.2020.00053.

[29] "'Why Should I Trust You?' | Proceedings of the 22nd ACM SIGKDD International Conference on Knowledge Discovery and Data Mining." https://dl.acm.org/doi/abs/10.1145/2939672.2939778 (accessed Dec. 16, 2022).

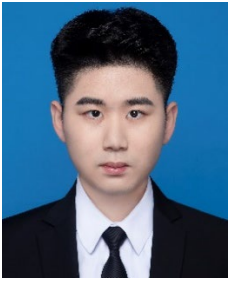

**ZHIBO ZHANG** received the Bachelor of Science degree in mechatronics engineering from Northwestern Polytechnical University, China, in 2021. He is currently pursuing a master's degree in electrical and computer engineering at Khalifa University, United Arab Emirates. He was awarded the Award for Outstanding Young Researcher by Khalifa University in 2022. His research interests focus on biometrics, cyber-physical system, cyber security, Explainable Artificial Intelligence, and Federated Learning.

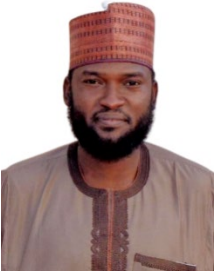

**SANI UMAR** received a Bachelor of Science (B.Sc) degree in Information Systems Engineering from Eastern Mediterranean University in the Turkish Republic of Northern Cyprus, and an M.Sc degree in Computer Networks from King Fahd University of Petroleum and Minerals, Kingdom of Saudi Arabia. He is currently pursuing a Ph.D. degree with the Department of Electrical Engineering and Computer Science, at Khalifa University of Science and Technology, Abu-Dhabi, United Arab Emirates. He received the best graduating student award during his B.Sc. His research interest includes Artificial Intelligence for cybersecurity (AI4SEC), Machine Learning (ML), Federated Learning (FL), and Mobile Crowd Sensing (MCS).

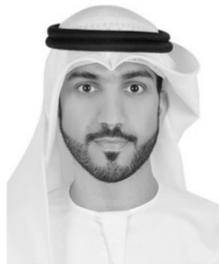

**AHMED Y. AL HAMMADI** received the B.S. degree in computer systems engineering from United Arab Emirates University (UAEU), UAE, and the M.Sc. degree in cryptology, security and coding of information systems from the Grenoble Institute of Technology, France. He is currently pursuing the Ph.D. degree with the Department of Electrical Engineering and Computer Science, Khalifa University of Science and Technology (KU), UAE. From 2007 to 2011, he worked for Global Communication and Software Systems as a Security Systems Expert and the acting Department Manager responsible for leading the public key infrastructure (PKI)-based security solution deployments in local governmental authorities and conducting continuous improvement research. From 2011 to 2020, he worked with the Department of Education and Knowledge as the Program Manager in the Special Projects Division of the Strategic Affairs Department to lead mainly different information technology initiatives that impact the entire education sector in Abu Dhabi. He was awarded the Best Paper Award in the 3rd International Conference on Electric Vehicle, Smart Grid, and Information Technology 2018 in South Korea. He also participated in the Abu Dhabi Government's efforts to digitalize the public services in the emirates and has been honored to participate in establishing a new governmental social entity in Abu Dhabi.

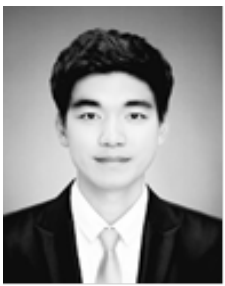

**SANGYOUNG YOON** holds M.Sc. (2019) degrees in Mechanical System Engineering from Chonbuk National University, Republic of Korea. After his M.Sc., he is currently a research associate at Electrical and Computer Engineering specializing in Artificial Intelligence in Khalifa University of Science and Technology, UAE. Previously worked as research associate at Civil Infrastructure and Environmental Engineering in Khalifa University. He is having involved in several research projects such as Cybersecurity, Deep Learning, and Optimization algorithms, Evaluation of porous media using Solitary wave, Impact damage analysis of carbon fiber reinforced composite. He is currently researching on non-destructive testing methods to detect composite defects using deep learning algorithms, as well as defenses against data

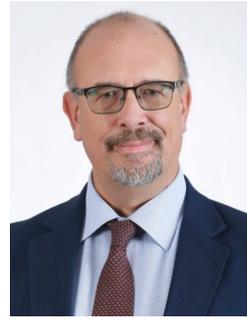

**ERNESTO DAMIANI** (Senior Member, IEEE) is currently a Full Professor with the Universitàdegli Studi di Milano, Italy, the Senior Director of the Robotics and Intelligent Systems Institute, and the Director of the Center for Cyber Physical Systems (C2PS), Khalifa University, United Arab Emirates. He is also the Leader of the Big Data Area, Etisalat British Telecom Innovation Center (EBTIC) and the President of the Consortium of Italian Computer Science Universities (CINI). He is also part of the ENISA Ad-Hoc Working Group on Artificial Intelligence Cybersecurity. He has pioneered model-driven data analytics. He has authored more than 650 Scopus-indexed publications and several patents. His research interests include cyber-physical systems, big data analytics, edge/cloud security and performance, artificial intelligence, and Machine Learning.

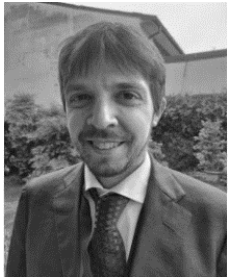

**Claudio Agostino Ardagna** is Full Professor at the Università degli Studi di Milano, the Director of the CINI National Lab on Big Data, and co-founder of Moon Cloud srl. His research interests are in the area of cloud-edge security and assurance, and data science. He has published more than 140 contributions in international journals, conference/workshop proceedings, and chapters in international books. He has been visiting researcher at BUTP, Khalifa University, GMU.

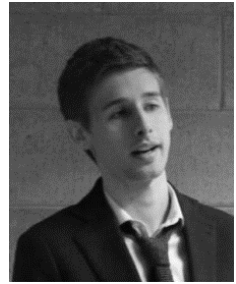

**Nicola Bena** is a Ph.D. student at the Università degli Studi di Mila-no. His research interests are in the area of security of modern dis-tributed systems with particular reference to certification, assurance, and risk management techniques

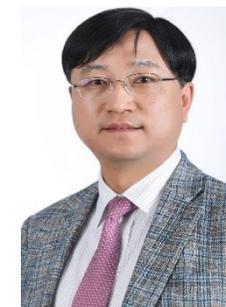

**CHAN YEOB YEUN** (Senior Member, IEEE) received the M.Sc. and Ph.D. degrees in information security from the Royal Holloway, University of London, in 1996 and 2000, respectively. After his Ph.D. degree, he joined Toshiba TRL, Bristol, U.K., and later became the Vice President at the Mobile Handset Research and Development Center, LG Electronics, Seoul, South Korea, in 2005. He was responsible for developing mobile TV technologies and related security. He left LG Electronics, in 2007, and joined ICU (merged with KAIST), South Korea, until August 2008, and then the Khalifa University of Science and Technology, in September 2008. He is currently a Researcher in cybersecurity, including the IoT/USN security, cyber-physical system security, cloud/fog security, and cryptographic techniques, an Associate Professor with the Department of Electrical Engineering and Computer Science, and the Cybersecurity Leader of the Center for Cyber-Physical Systems (C2PS). He also enjoys lecturing for M.Sc. cyber security and Ph.D. engineering courses at Khalifa University. He has published more than 140 journal articles and conference papers, nine book chapters, and ten international patent applications. He also works on the editorial board of multiple international journals and on the steering committee of international conferences.